%% file: main.tex
\pdfoutput=1

\documentclass[11pt]{article}

\usepackage[final]{acl}

\usepackage{todonotes}
\usepackage{xcolor}
\usepackage{fancyvrb}
\usepackage{fvextra}

\usepackage{amssymb}

\usepackage{CJKutf8}

\usepackage[breakable]{tcolorbox}

\usepackage{fancyvrb}
\DefineVerbatimEnvironment{Verbatim}{Verbatim}{fontfamily=cmr}
\definecolor{LightTeal}{RGB}{168, 216, 216}  %
\definecolor{LightGold}{RGB}{255, 210, 157}  %

\usepackage{times}
\usepackage{latexsym}

\usepackage[T1]{fontenc}
\usepackage[utf8]{inputenc}

\usepackage{microtype}

\usepackage{inconsolata}

\usepackage{graphicx}

\usepackage{acro}

\usepackage{caption}

\usepackage{booktabs}

\title{Same evaluation, more tokens: On the effect of input length for machine translation evaluation using Large Language Models}

\author{%
  Tobias Domhan\thanks{Now at Google. Correspondence to: \texttt{domhant@google.com}.}  \and Dawei Zhu\\
  Amazon AGI\\
  \texttt{\{domhant,daweizhu\}@amazon.com}
}

\DeclareAcronym{LLM}{
short=LLM,
long=Large Language Model
}

\DeclareAcronym{MQM}{
short=MQM,
long=Multidimensional Quality Metrics
}

\DeclareAcronym{CoT}{
short=CoT,
long=Chain-of-Thought
}

\DeclareAcronym{ICL}{
short=ICL,
long=in-context learning
}

\begin{document}

\maketitle
\begin{abstract}

Accurately evaluating machine-translated text remains a long-standing challenge, particularly for long documents. Recent work has shown that large language models (LLMs) can serve as reliable and interpretable sentence-level translation evaluators via MQM error span annotations. With modern LLMs supporting larger context windows, a natural question arises: can we feed entire document translations into an LLM for quality assessment? Ideally, evaluation should be invariant to text length, producing consistent error spans regardless of input granularity. However, our analysis shows that text length significantly impacts evaluation: longer texts lead to fewer error spans and reduced system ranking accuracy. To address this limitation, we evaluate several strategies, including granularity-aligned prompting, Focus Sentence Prompting (FSP), and a fine-tuning approach to better align LLMs with the evaluation task. The latter two methods largely mitigate this length bias, making LLMs more reliable for long-form translation evaluation.

\end{abstract}

\section{Introduction}

Historically, the field of Machine Translation has been dominated by a sentence-level paradigm, where individual sentences are translated in isolation. \acfp{LLM}, with increasingly long context windows, are able to process hundreds of thousands of tokens of context~\cite{achiam2023gpt}.
With the right set of prompts, they can be used to translate full documents~\cite{wu2024adapting,briakou-etal-2024-translating}, potentially moving beyond the sentence-level paradigm. As machine translation expands to longer texts, a key challenge is developing reliable methods for automatic evaluation.
Trained translation metrics have been shown to be able to evaluate long text spans~\cite{vernikos-etal-2022-embarrassingly, raunak2023slide}, despite being trained on sentence-level data.
However, their application to longer texts is often constrained by the context window limitations of their base models, e.g., a 512-token limit~\cite{conneau-etal-2020-unsupervised}.
At the same time, \acp{LLM} have demonstrated SOTA performance in evaluating short-form translations~\cite{freitag-etal-2023-results, freitag-etal-2024-llms, kocmi-federmann-2023-gemba}, when prompted to produce error spans with categories defined by \ac{MQM} \cite{lommel2014multidimensional}.
This raises the question: \textit{can we feed increasingly longer translations into LLMs to achieve reliable long-form translation evaluation?}
In this work, we define short-form to refer to single or few sentences, while long-form refers to longer units of text, such as multiple paragraphs or documents.

\begin{figure}[t]
  \includegraphics[width=\columnwidth]{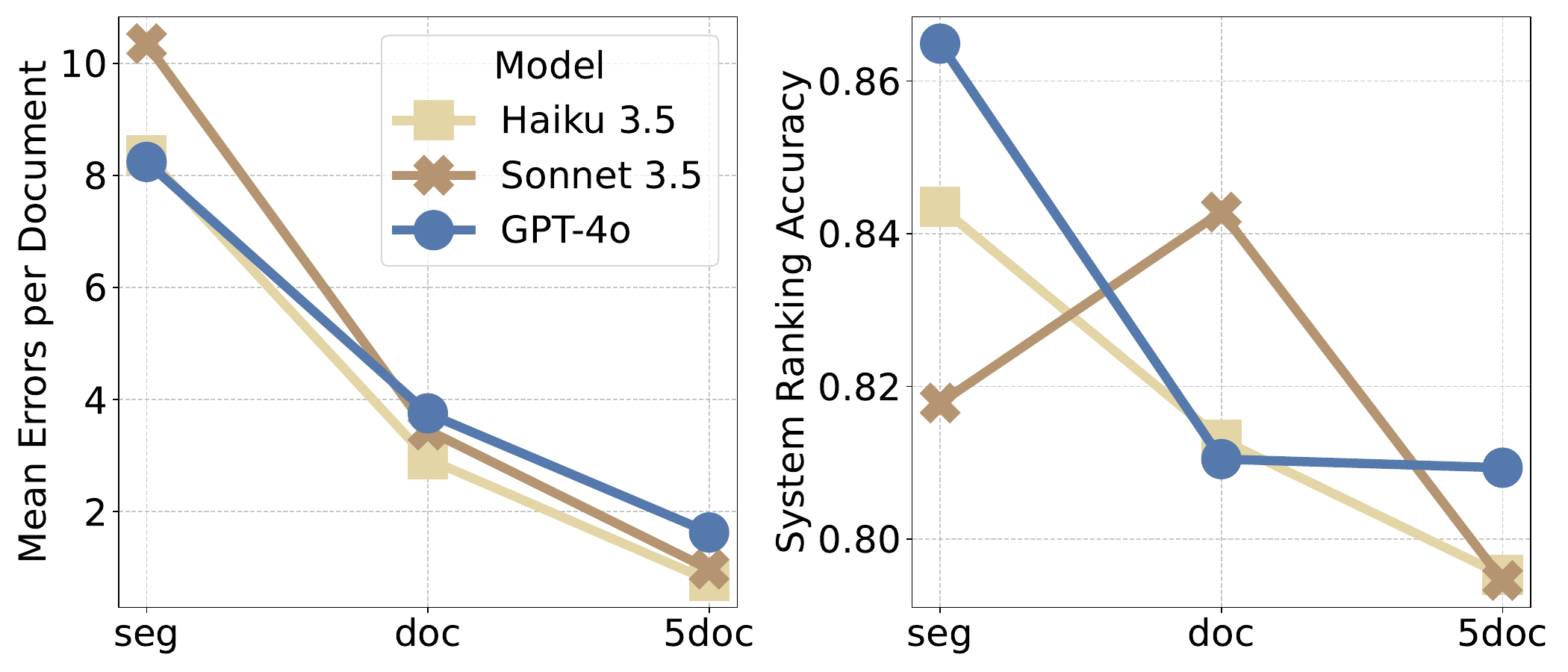}
  \caption{The number of predicted errors (left) and the average accuracy (right) for different input text granularities: segment-level, doc-level, and 5 doc-level.
  }
  \label{fig:numspans-vs-accuracy}
\end{figure}

Ideally, when providing a long document translation for evaluation, LLMs should thoroughly process all sentences and flag all errors present.
However, we find that current LLMs are not ``length-invariant'': they detect significantly fewer errors when assessing an entire document at once, compared to the cumulative number of errors identified when processing the document one segment at a time (Figure~\ref{fig:numspans-vs-accuracy}, left). In other words, many errors are missed when evaluating long documents.
Having fewer errors identified reduces the interpretability.
Even worse, we find that for Claude 3.5 Haiku and GPT-4o, the average ranking accuracy also decreases as the input length increases (Figure~\ref{fig:numspans-vs-accuracy}, right).
This aligns with prior work showing that \acp{LLM} struggle with reasoning tasks as the input length increases~\cite{levy-etal-2024-task}.
We argue that evaluating long-form translations with LLMs requires more refined approaches. Our core contributions are: \textbf{(1)} a comprehensive analysis of length dependence in current LLMs; \textbf{(2)} a prompting scheme that ensures stable ranking accuracy and error detection across text granularities; and \textbf{(3)} practical guidelines for deploying LLMs in long-form translation evaluation, informed by results from diverse models and settings.
\section{Length-invariant translation evaluation}

\acp{LLM} are shown to be competitive with SOTA translation metrics when used to predict \ac{MQM} error spans~\cite{fernandes-etal-2023-devil, kocmi-federmann-2023-gemba, freitag-etal-2023-results,freitag-etal-2024-llms}. For example, \citet{kocmi-federmann-2023-gemba} propose the GEMBA-MQM prompt to instruct GPT-4 to predict translation error spans, along with their severities. Error weights associated with severities are then summed at the segment or system level to produce a final quality score. However, we find that LLMs become less reliable when evaluating long-form translation outputs (Figure~\ref{fig:numspans-vs-accuracy}). This could be because long responses are less commonly encountered in NLP tasks.

\begin{figure}[th]
    \centering
    \includegraphics[width=\columnwidth]{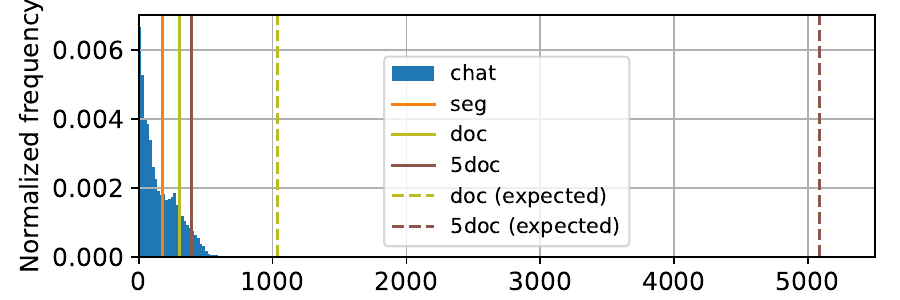}
    \caption{Comparison of chat response lengths~(\textit{chat}) compared to Claude 3.5 Sonnet's MQM response length (\textit{seg}, \textit{doc}, \textit{5doc}) on WMT'24 EN-DE metrics task data in GPT-4o tokens. The expected length is based on the concatenation of the segment level responses. }
    \label{fig:output-lengths}
\end{figure}

In Figure~\ref{fig:output-lengths}, we contrast the response length of typical user interactions with chat systems, based on Chatbot Arena~\cite{zheng2023judging}, to the expected response length of Claude for MQM annotation of different text lengths.\footnote{Three levels of text lengths are considered: \textit{seg}, \textit{doc}, \textit{5doc}, with increased length. Refer to Section~\ref{sec:setup} for the definition.} We observe responses to longer inputs to be substantially shorter than responses concatenated from segment-level responses. For instance, asking Claude to flag all MQM errors in document-level translations yields an average of around 300 tokens (solid green line). However, if we split the same documents into segments, flag errors at the segment level,\footnote{In WMT’24, parallel data is segment-level with metadata linking segments from the same document, enabling both segment- and document-level evaluation (via concatenation).} and then concatenate all flagged errors, we obtain an average of \textasciitilde1,000 tokens (dashed green line). In fact, 99\% of general chat responses are shorter than 516 tokens, and response length required to cover all error spans in a document clearly falls outside this range. In this work, our goal is to develop an evaluation scheme that enables \acp{LLM} to reliably assess long-form translation, leading to consistent results across input granularities. To this end, we explore several prompting and fine-tuning strategies.
\begin{minipage}{0.9\columnwidth}%
\makebox[\columnwidth]{%
  \includegraphics[width=1.0\columnwidth]{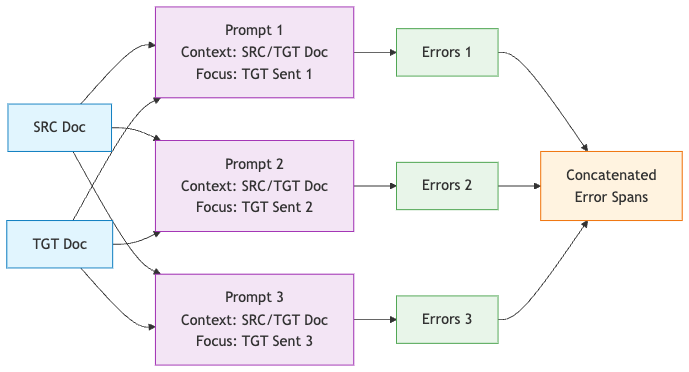} }
\captionof{figure}{FSP on a three-sentence document.}\label{fig:fsp}
\end{minipage}

\paragraph{Focus sentence prompting (FSP)}
To ensure invariance to input length, we consider individual sentences as the evaluation unit for MQM error span prediction. Specifically, we present the evaluation model with the full source and target documents, but prompt it to evaluate a single sentence at a time. We call this approach Focus Sentence Prompting (FSP). Here, we emphasize that we consider a realistic long-form translation scenario where the entire document is translated in a single pass, which better accommodates discourse phenomena~\cite{maruf-docmt-19}. Therefore, the translation may not strictly follow a one-to-one correspondence between source and target sentences. FSP effectively avoids the need for sentence alignment, a process that often introduces noise from wrong alignments. Moreover, providing the full source and translation text enables the model to detect context-dependent errors, such as those related to anaphora resolution. While FSP, by design, requires multiple inference passes, resulting in increased inference cost, this overhead can be mitigated by prompt caching, as all prompts for a document share the same prefix. Figure~\ref{fig:fsp} illustrates the FSP schema, and the full prompt is provided in Appendix~\ref{prompt:tgtsent}.

\paragraph{Granularity Matching} The original GEMBA-MQM prompt uses a fixed set of three sentence-level demonstrations for MQM annotations, which can lead to significant mismatches in text granularity when the test case requires long-form translation evaluation, potentially causing the model to generate fewer error spans. To address this, we use two approaches. One method retains \ac{ICL} but selects five demonstrations that roughly match the length of the test example (\textbf{GMICL-5}). The other method fine-tunes LLMs on MQM data, similar to \citet{fernandes-etal-2023-devil}, but at various text granularities, referred to as \textbf{GMFT}. The data source for both GMICL-5 and GMFT comes from the WMT'23 shared task data~\cite{freitag-etal-2023-results}.

\paragraph{Error Span Explanations}
Inspired by Chain-of-Thought~\cite{neurips-2022-cot}, our MQM prompts (e.g., FSP and GMICL-5) ask LLMs to predict both error spans and their corresponding explanations unless stated otherwise.
The error category and severity are then predicted based on the span-level explanation, aiming for more accurate judgment. In Appendix~\ref{sec:error_span_ablation}, we show that adding explanations leads to higher system ranking accuracy.

\paragraph{Direct Assessment (DA)} Instead of deriving translation quality scores from predefined error weights, LLMs can be prompted to predict a numerical score for quality assessment based on the identified error spans, a method referred to as Direct Assessment (DA)~\cite{kocmi-federmann-2023-large}.

\section{Experiments}
\label{sec:experiments}

\begin{figure*}[th]
    \centering
    \includegraphics[width=2\columnwidth]{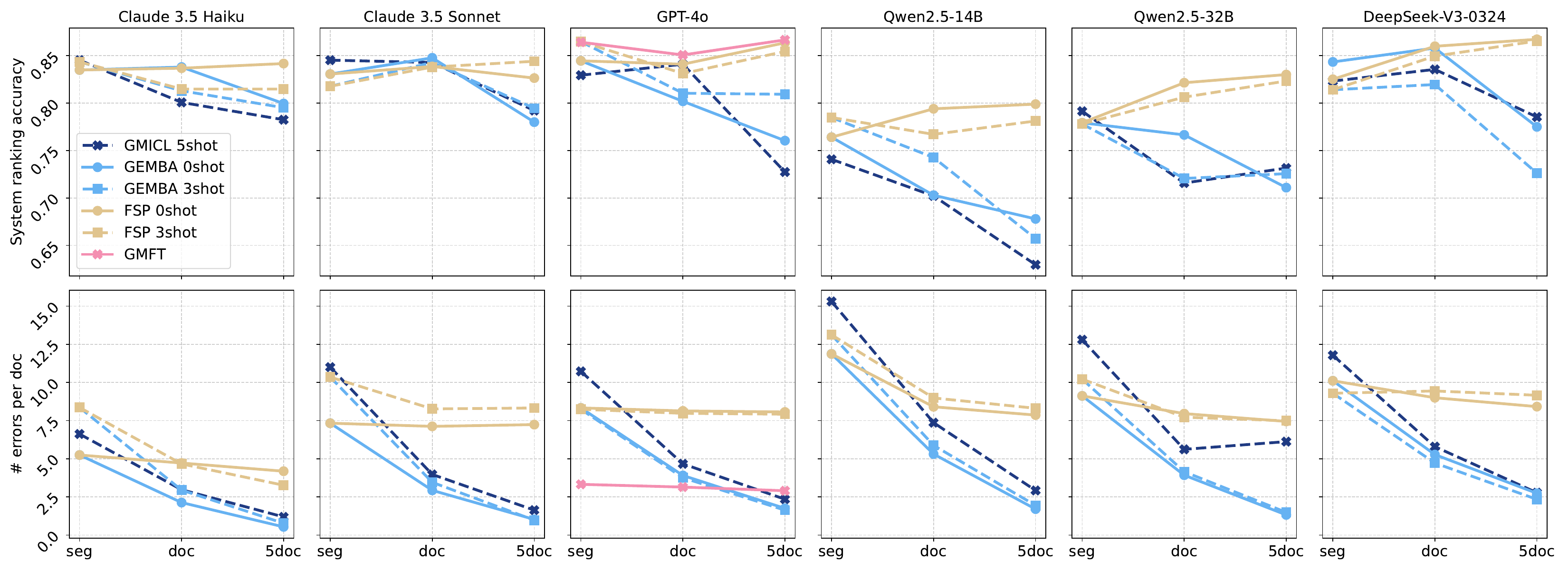}
    \caption{System ranking accuracy and number of error spans per document across methods and input granularities, averaged across translation directions. See Appendix~\ref{sec:translation_direction_specific_results}, \ref{sec:precision_recall_f1} for direction-specific results and character F1 scores.}
    \label{fig:main-results}
\end{figure*}

\subsection{Setup}
\label{sec:setup}

We use the WMT'24 metrics shared task data for evaluation~\cite{freitag-etal-2024-llms}.
The data consists of human gold \ac{MQM} annotations covering three translation directions: English $\rightarrow$ German (EN-DE), Japanese $\rightarrow$ Chinese (JA-ZH) and English $\rightarrow$ Spanish (EN-ES). We use system-level pairwise accuracy ~\cite{kocmi-etal-2021-ship} as our evaluation metric. It measures the number of pairs of systems that are ranked correctly when compared to the ranking derived from human annotations.
We use the official shared task scripts to access data and compute metrics.\footnote{\href{https://github.com/google-research/mt-metrics-eval}{github.com/google-research/mt-metrics-eval}}
Additionally, we measure the number of error spans per document and the character F1 score.
The latter is also used by the WMT shared task on quality estimation~\cite{blain2023findings} and based on the precision/recall of error spans compared to gold annotations per character with partial credit (0.5) for a mismatch in error severity (more details in Appendix~\ref{sec:precision_recall_f1}). We evaluate three proprietary \acp{LLM}: Claude 3.5 Haiku, Claude 3.5 Sonnet, and GPT 4o, as well as three open weight \acp{LLM}: Qwen2.5 14B/32B~\cite{qwen2025qwen25}, and DeepSeek V3~\cite{deepseekai2025deepseekv3}, using a temperature of 0 for deterministic decoding.
The WMT'24 metrics shared task data is provided at the segment level, with each segment containing one or a few sentences. These segments originate from documents, which we reconstructed using the provided metadata. Additionally, we concatenate random groups of five documents to simulate extended long-form input. This results in three evaluation settings with different text granularities, referred to as \textit{seg}, \textit{doc}, and \textit{5doc}. The gold \ac{MQM} annotations for \textit{doc} and \textit{5doc} are derived by concatenating the \ac{MQM} errors from the segment-level annotations. This leads to an average of 103/507/2713 GPT-4o tokens for the \textit{seg}, \textit{doc} and \textit{5doc} cases. Similarly, we sample text data from WMT'23 with the three granularity levels for demonstrations and training data for GMICL-5 and GMFT.
Refer to Appendix~\ref{sec:gm-data-construction} for data construction details.
For GMFT, we fine-tune a GPT-4o model.

\subsection{Addressing evaluation length dependence}
Figure~\ref{fig:main-results} shows the results of the evaluated prompt and fine-tuning settings across different text granularities.
Compared to the GEMBA 3-shot baseline, which uses a fixed set of three segment-level MQM demonstrations, the GMICL-5 approach, despite being designed to match the input granularity, results in only marginal increases in error spans and still suffers a substantial drop in system ranking accuracy for longer texts. This suggests that \textbf{providing additional test-like demonstrations alone is not sufficient to overcome the response length bias (Figure~\ref{fig:output-lengths}) or to improve accuracy.}

In contrast, FSP effectively stabilizes the number of errors across text granularities and models. It also improves system ranking accuracy, particularly for moderate-sized LLMs and long-document scenarios (e.g., +12\% accuracy with Qwen2.5-14B on 5doc compared to GEMBA), suggesting that LLMs can accurately identify the relevant source context for evaluating translation segments, even without explicit alignment.
\textbf{Overall, FSP, when paired with strong LLM judges, can serve as a reliable, long-context, reference-free evaluation metric.}
For example, FSP with GPT-4o ranks second among official WMT’24 submissions (full ranking in Appendix~\ref{sec:comparison_wmt24}). Finally, we examine the impact of shot count in GEMBA and FSP on accuracy: no consistent pattern emerges in favor of 3-shot over 0-shot. The most effective setting depends on the specific \acp{LLM} and text granularity.

Our fine-tuning experiments with GPT-4o (GMFT) further demonstrate that a small amount of training data may also mitigate length bias. This results in a more consistent error count across text granularities while maintaining high system ranking accuracy, serving as an alternative to FSP.
Interestingly, despite the overall stability, the predicted error count is lower at the \textit{seg} and \textit{doc} levels compared to other methods, suggesting that a high number of error spans may not be essential for higher ranking accuracy.
The number of errors predicted by GMFT depend on the number of errors in the human annotation training data, which contains fewer, higher quality error spans.
We therefore focus on comparing the number of error spans at different granularities for a given model, not comparing across models at a specific granularity.

\begin{figure}[th]
    \centering
    \includegraphics[width=\columnwidth]{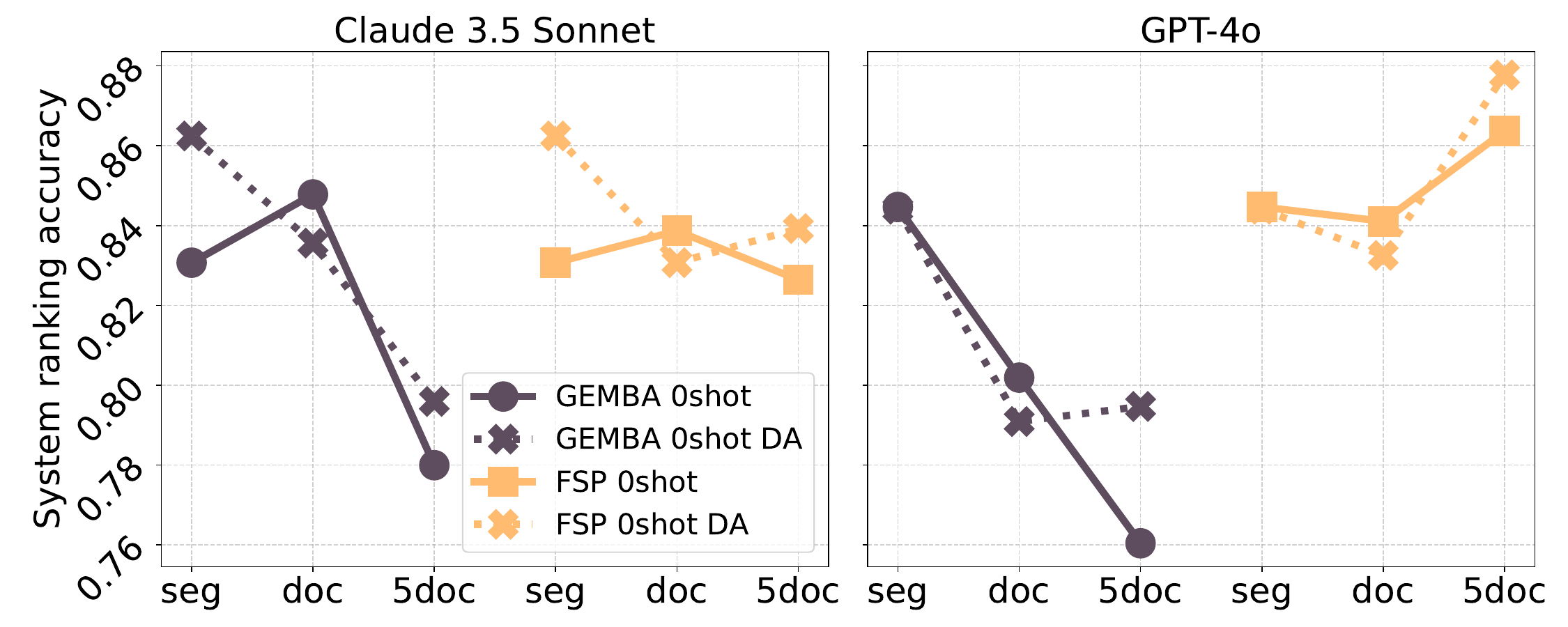}
    \caption{Impact of predicting a quality score via DA compared to computing a weighted MQM error score for GEMBA and FSP prompting.}
    \label{fig:da-comparison}
\end{figure}

Additionally, we explore whether supplementing GEMBA with DA could mitigate length dependence.
In the DA setting a single quality score is predicted, rather than just identifying errors.
Figure~\ref{fig:da-comparison} reveals that DA, by itself, is still vulnerable to text length variations and does not improve ranking accuracy for long-form translations. Nevertheless, DA integrates well with FSP, likely because the finer-grained error spans provided by FSP help LLMs infer a more accurate quality score.

\subsection{FSP inference efficiency}

FSP increases input tokens by adding document context to each segment, raising concerns about inference efficiency.  
In Table~\ref{tab:fsp_efficiency}, we compare the throughput in terms of error spans per second.
Despite the increased number of input tokens for FSP, we can see that the throughput is comparable between FSP and GEMBA 3shot.
The reason is that for autoregressive models inference time is dominated by the output token generation, whereas FSP only increases the number of input tokens.
Additionally, prompt caching, as available in inference frameworks like SGLang~\cite{zheng2024sglang}, is effective for FSP as all evaluation prompts for a given document share the same prefix.
Note that the inference time is shorter for GEMBA 3shot due to producing fewer error spans at a lower accuracy.

\begin{table}[ht]
  \centering
  \resizebox{\columnwidth}{!}{%
    \begin{tabular}{@{}l c c c  c c c@{}}
      \toprule
        & \multicolumn{3}{c}{doc} 
        & \multicolumn{3}{c}{5doc} \\
      \cmidrule(lr){2-4}\cmidrule(lr){5-7}
        & duration & \# spans/s & Acc. 
        & duration & \# spans/s & Acc. \\
      \midrule
      \multicolumn{7}{c}{\textit{Qwen2.5-32B}} \\
      GEMBA 3shot  & 29min & 36.6 & 72.1 & 13min & 29.1 & 72.6 \\
      FSP 3shot & 58min & 34.4 & \textbf{80.6} & 58min & 33.2 & \textbf{82.3} \\
      \multicolumn{7}{c}{\textit{DeepSeek-V3-0324}} \\
      GEMBA 3shot  & 63min & 19.4 & 82.0 & 46min & 13.3 & 72.6 \\
      FSP 3shot    & 151min & 16.3 & \textbf{85.0}  & 156min & 15.2 & \textbf{86.6} \\
      \bottomrule
    \end{tabular}%
  }
\caption{Inference performance comparison. Models are deployed with SGLang. Duration denotes wall-clock time to complete the WMT’24 metrics shared task.}
\label{tab:fsp_efficiency}
\end{table}

\section{Related work}

\citet{kocmi-federmann-2023-gemba} and \citet{fernandes-etal-2023-devil} show that \acp{LLM} can be effective translation evaluators when prompted to predict \ac{MQM} error spans.
Due to the lack of public document-level MQM data for meta-evaluation, their evaluation focuses on sentence-level translations.\footnote{With the exception of WMT23 EN-DE being paragraph-level data.}
We overcome this data limitation by combining the data into longer text blocks comprising single or multiple documents.
\citet{fernandes-etal-2023-devil} demonstrate that \acp{LLM} can be effectively fine-tuned for the MQM error-span prediction task.
In the GMFT approach we fine-tune a model on the error span task, extending the setup to texts of different lengths.

For trained, dedicated translation metrics that predict quality scores without fine-grained errors or explanations, it has been shown that sentence-level metrics can be extended to evaluate longer texts~\cite{raunak2023slide,vernikos-etal-2022-embarrassingly}.
xCOMET~\cite{guerreiro-etal-2024-xcomet} is a recent encoder-only model fine-tuned to be able to predict both a quality score and per-token error severities.
As a fine-tuned model, it comes with the downside of not being able to use the latest LLMs, as well as not providing error explanations.
The second downside was later overcome by xTower~\cite{treviso-etal-2024-xtower}, showing that error explanations are useful for error correction.
We include error-span explanations, as we find that they improve the ranking accuracy.

\section{Conclusion}
In this work, we show that SOTA LLMs lack length invariance in translation assessment, detecting fewer error spans at the document level than when evaluating segments individually.
The system ranking accuracy also drops with longer inputs.
We observed this behavior across a range of open and closed models.
To address this, we propose two simple yet effective methods for length-invariant, accurate evaluation.

\section*{Limitations}
To ensure transparency and foster future research, we outline several limitations of our study below.

\paragraph{Limited Translation Directions}
Due to limitations in the publicly available MQM datasets that include document-level metadata, we were able to run experiments on only three language directions.
As more MQM datasets become available, we encourage other researchers to replicate our experiments to see whether the findings hold for other language directions. 
While our work was limited to these three language directions, we have no strong reason to believe that our findings will not generalize to other language combinations.

\paragraph{Evaluation Scope} We evaluated the impact of length on long-form translation assessment using the MQM metric. While MQM is a widely accepted standard, we did not extend it to explicitly address document-level phenomena such as anaphora resolution, coherence, or consistency across a document. However, we note the following: (1) achieving length invariance in long-form evaluation is a critical prerequisite that must be addressed before focusing on other important aspects of document-level translation quality; and (2) extending standard metrics falls outside the scope of this paper. Nonetheless, we consider this a promising direction for future research. In particular, future work could involve designing experiments to evaluate the ability of LLMs to identify and assess document-level translation errors, which may require annotated corpora.

\paragraph{Sentence Segmentation Requirement} Our work assumes the availability of segment-level data. In practical applications, sentence segmentation would be necessary. However, this should not pose a significant challenge, as sentence segmentation tools are readily available.
This assumption allows us to focus on the core issue of the lack of length invariance, which we see as a prerequisite for long-form translation evaluation.
Note, that FSP would not able to penalize omissions of entire sentences, which is why advocate GMFT where available.
Duplicate sentences, which are rare in practice, might requiring including context or assigning unique identifiers for FSP.

\section*{Acknowledgments}
We thank Bill Byrne, Felix Hieber, Michael Denkowski, and Raúl Soutelo Quintela for their in-depth discussions and valuable feedback that helped shape this research. We would also like to thank our anonymous reviewers for their constructive feedback.

\bibliography{custom}

\clearpage
\appendix

\section{Experiment details}
\label{sec:appendix}

\subsection{LLM versions}
The experiments are based on these versions of Claude, GPT, and DeepSeek V3:
\begin{itemize}
    \item Sonnet 3.5:\\ \textit{anthropic.claude-3-5-sonnet-20241022-v2:0}
    \item Haiku 3.5:\\\textit{anthropic.claude-3-5-haiku-20241022-v1:0}
    \item GPT-4o:\\ \textit{gpt-4o-2024-11-20}
    \item DeepSeek V3:\\ \textit{DeepSeek-V3-0324}
\end{itemize}

\section{Inference infrastructure}
We utilize the official APIs to prompt Claude and GPT models. Qwen and DeepSeek models are deployed through SGLang~\cite{zheng2024sglang} on NVIDIA H200 GPUs.

\subsection{Data statistics}

Table~\ref{tab:data-stats} contains the number of segments/documents and multi-documents as well as their average length in GPT-4o tokens both per language arc and the average across directions. The number of tokens is computed via the \texttt{tiktoken} library\footnote{\url{https://github.com/openai/tiktoken}}.

\begin{table*}[ht]
\centering
\begin{tabular}{|l|r|r|r|r|r|r|r|r|}
\hline
  & \multicolumn{2}{c|}{\textbf{EN-DE}} & \multicolumn{2}{c|}{\textbf{JA-ZH}} & \multicolumn{2}{c|}{\textbf{EN-ES}} & \multicolumn{2}{c|}{\textbf{average}} \\
\textbf{Granularity} & \textbf{\# items} & \textbf{\# tokens} & \textbf{\# items} & \textbf{\# tokens} & \textbf{\# items} & \textbf{\# tokens} & \textbf{\# items} & \textbf{\# tokens} \\
\hline
\textit{seg} & 27,944 & 90.4 & 16,606 & 129.1 & 23,952 & 88.0 & 22,834 & 102.5 \\
\textit{doc} & 4,788 & 530.1 &  4,531 & 474.4 &  4,104 & 516.3 & 4,474 & 506.9 \\
\textit{5doc} & 980 & 2814.2 & 920 & 2576.4 &  840 & 2747.0 & 913.3 & 2712.5 \\
\hline
\end{tabular}
\caption{WMT'24 metrics shared task evaluation data statistics. The number of tokens is the average number of source and translation text GPT-4o tokens at the given granularity.}
\label{tab:data-stats}
\end{table*}

\subsection{Error span explanation ablation}
\label{sec:error_span_ablation}

\begin{figure}[th]
    \centering
    \includegraphics[width=0.9\columnwidth]{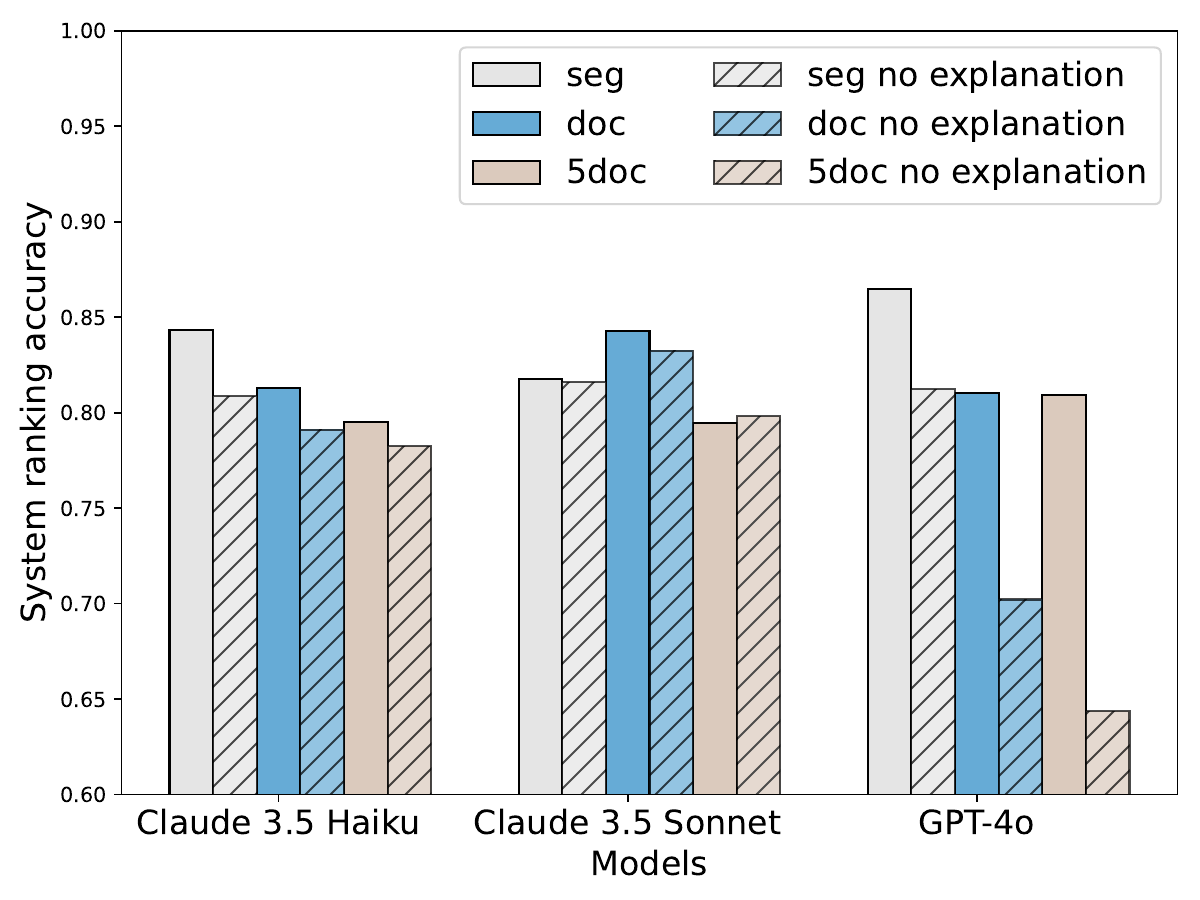}
    \caption{Ablation comparing the GEMBA 3-shot prompt variation with and without error span explanations.}
    \label{fig:explain-vs-noexplain}
\end{figure}

Figure~\ref{fig:explain-vs-noexplain} compares the GEMBA 3-shot prompt variant once with and once without error span explanations.
We see that for both Claude 3.5 Haiku and GPT-4o error span explanations significantly improve system ranking accuracy, while showing the same pattern of decreasing accuracy with  longer inputs.
For GPT-4o the decrease in system ranking accuracy is even more pronounced when not using error span explanations.
For Claude 3.5 Sonnet the accuracy is comparable between the variant with and without error span explanations. 
The ranking of what works best in terms of granularities is unchanged though.

\subsection{Granularity-Matching Data Construction}
\label{sec:gm-data-construction}

We constructed a dataset with MQM annotations at various text granularities. Our data is derived from the WMT23 MQM dataset~\cite{freitag-etal-2023-results}, which was originally provided at the segment level for both source texts and machine translations. Similar to the approach used to create the WMT’24-based test data described in Section~\ref{sec:experiments}, we joined segments from the same documents to form document-level examples. Subsequently, we concatenated five randomly selected documents from these document-level examples to create \textit{5doc} long examples. In total, our dataset comprises 43307 examples, including 35472, 6530, and 1305 examples for the granularities \textit{seg}, \textit{doc}, and \textit{5doc}, respectively. These examples cover the translation directions English-German, Hebrew-English, and Chinese-English.

For GMICL-5, we randomly select five granularity-matched examples from our constructed dataset. For fine-tuning GPT-4o, we use all 43307 examples and fine-tuned the GPT-4o models for two epochs through OpenAI's API. It is important to note that the translation directions in training and testing differ, with only English-German overlapping. However, we find that fine-tuning still improved GPT-4o’s length invariance across all tested directions.

The statistics for the resulting data at different granularities can be found in Table~\ref{tab:data-stats}.

\subsection{Translation Direction-Specific Results}
\label{sec:translation_direction_specific_results}
\begin{figure*}[th]
    \centering
    \includegraphics[width=2\columnwidth]{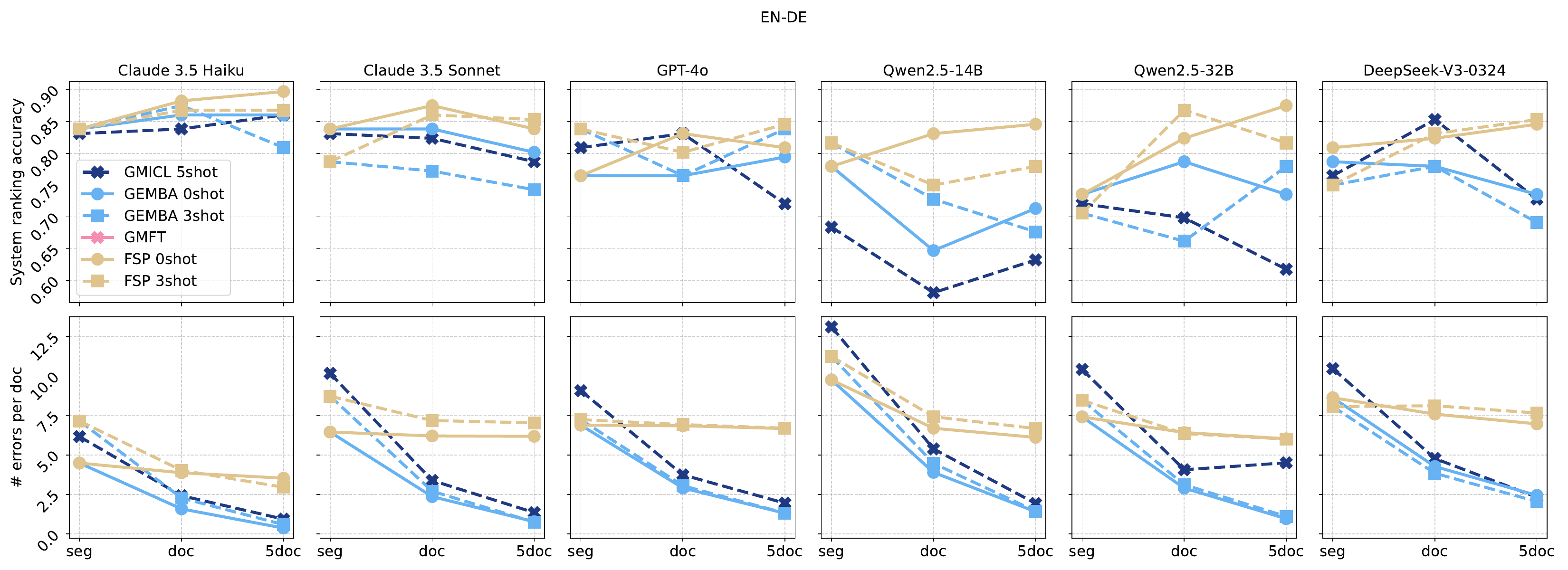}
    \includegraphics[width=2\columnwidth]{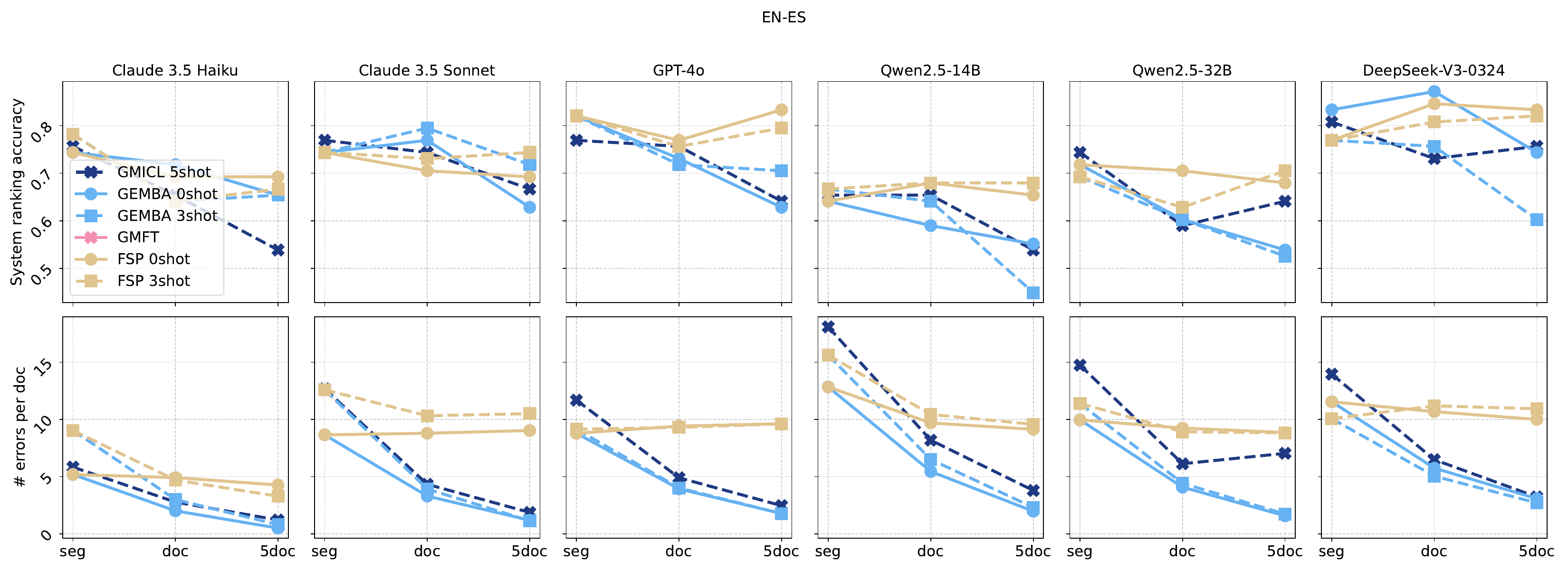}
    \includegraphics[width=2\columnwidth]{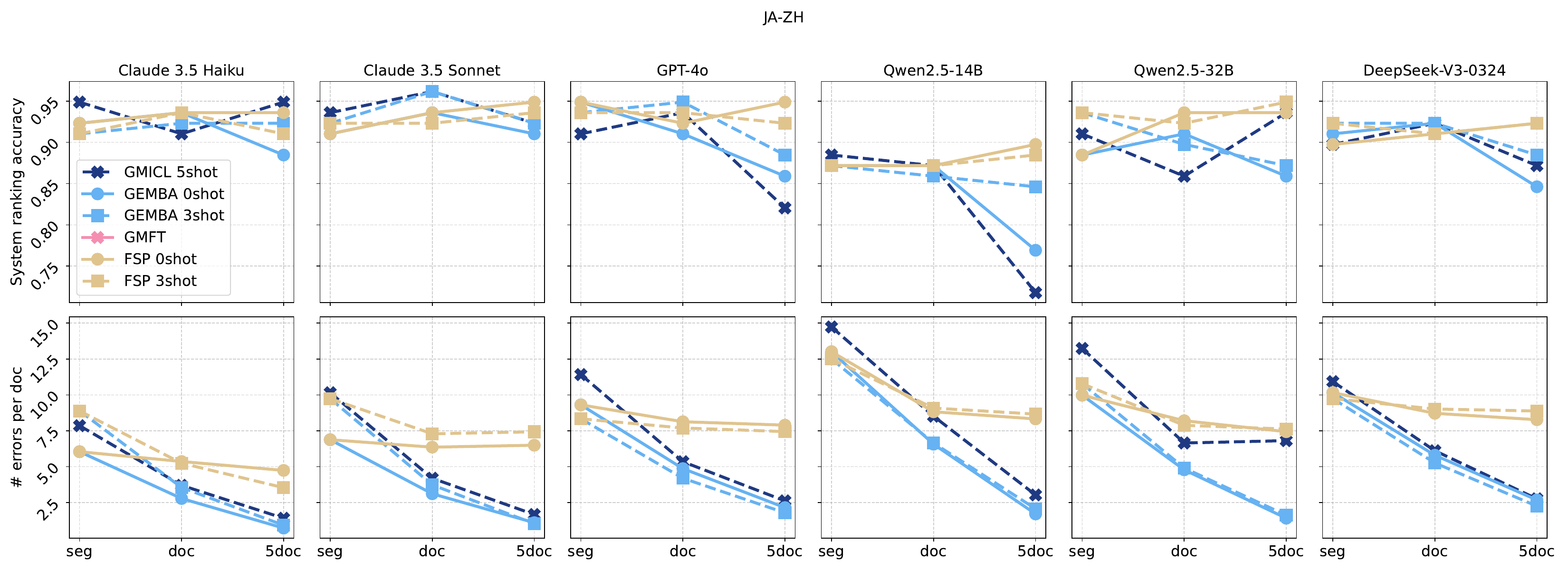}
    \caption{System ranking accuracy and number of error spans at different input granularities for individual translation directions: EN-DE, EN-ES, and JA-ZH.}
    \label{fig:main-results-direction-individual}
\end{figure*}

In Figure~\ref{fig:main-results-direction-individual}, we show the system ranking accuracy, number of error spans, and F1 score at various input granularities for individual translation directions: EN-DE, EN-ES, and JA-ZH. The results demonstrate that FSP generally outperforms other approaches across different translation directions and models, particularly in the 5doc case.

\subsection{Character-level precision/recall/F1}
\label{sec:precision_recall_f1}
\paragraph{Implementation Details}In order to compute character-level precision/recall/F1 we need to know the location of error spans. The GEMBA-MQM prompts however only result in error span strings without a specific location.
While the majority of error spans is unique in the translation string (86.37\% of Haiku 3.5's error spans using the FSP prompt on EN-DE data at the \textit{doc5} granularity) there are shorter spans that occur multiple times.
To assign a location for them we greedily search for all possible locations and pick the one that is unoccupied by any other error span choosing the one with the highest gold annotation overlap.
This optimistically assumes the model refers to the correct location. As this is done equally for all models and systems this does not give an unfair advantage.
The chosen location is marked as occupied for each character it spans and we move to the next error span.
There may be error spans that can not be matched to a location, which will reduce the precision, while not affecting the recall.
Once all error spans have been assigned to locations we can proceed to compute the character-level precision and recall.
The precision computes the number of correctly predicted characters, while recall computes the number of gold characters that the model has covered.

\begin{figure*}[th]
    \centering
\includegraphics[width=2\columnwidth]{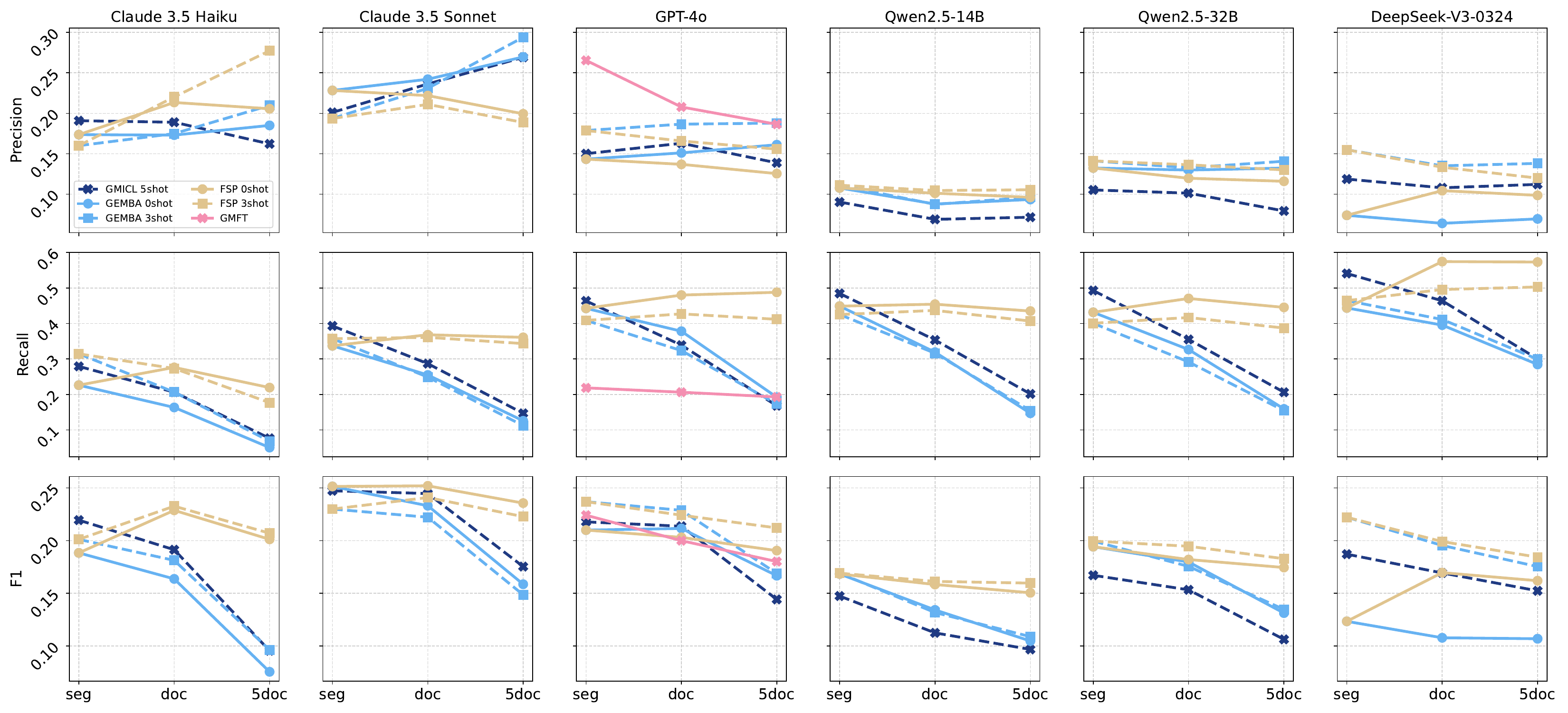}
    \caption{Character precision, recall and F1, averaged across translation directions.}
    \label{fig:appendix-precision-recall}
\end{figure*}

\paragraph{Results} Figure~\ref{fig:appendix-precision-recall} contains the character-level precision, recall, and F1 under different evaluation setups. As can be seen, FSP not only results in a higher and more stable error distribution but also improves the overlap with gold spans, measured using character F1, when evaluating long documents. This suggests that the error spans predicted by the model more closely align with those identified by human annotators.

\subsection{Comparison to WMT'24 shared task submissions}
\label{sec:comparison_wmt24}
Our goal is to provide an evaluation setting that allows using off-the-shelf LLMs for long-form translation evaluation, not to produce the state-of-the-art in segment-level translation metrics.
Nevertheless to see how our prompting setup compares to WMT'24 metrics task submissions we included results in terms of system ranking accuracy in Table~\ref{tab:comparison-to-wmt24-submissions}.
We see that, especially when using GPT-4o, our evaluation settings, that includes JSON outputs and error explanations, are competitive with state-of-the-art metrics at the segment level.
Note, that WMT'24 moved from system ranking accuracy to a soft variant that takes the uncertainty into account, Soft Pairwise Accuracy (SPA)~\cite{thompson-etal-2024-improving}. 
We chose to continue using system ranking accuracy as SPA requires segment-level scores, which makes it more challenging to compare the same evaluations across different text granularities as we do in this work.

\begin{table*}[ht]
\centering
\begin{tabular}{lcr}
\textbf{System} & \textbf{ours?} & \textbf{system ranking accuracy} \\
\verb|*CometKiwi-XXL[noref]| & & 0.8692 \\
\verb|GPT-4o FSP 3shot[noref]| & $\checkmark$ & 0.8649 \\
\verb|*MetricX-24| & & 0.8655 \\
\verb|MetricX-24-Hybrid| & & 0.8594 \\
\verb|gemba_esa[noref]| & & 0.8520 \\
\verb|*metametrics_mt_mqm_same_source_targ| & & 0.8485 \\
\verb|metametrics_mt_mqm_hybrid_kendall| & & 0.8485 \\
\verb|XCOMET| & & 0.8472 \\
\verb|*metametrics_mt_mqm_kendall| & & 0.8460 \\
\verb|*MetricX-24-QE[noref]| & & 0.8454 \\
\verb|Claude 3.5 Haiku FSP 3shot[noref]| & $\checkmark$ & 0.8435 \\
\verb|MetricX-24-Hybrid-QE[noref]| & & 0.8338 \\
\verb|_BLEURT-20| & & 0.8210 \\
\verb|Claude 3.5 Sonnet FSP 3shot[noref]| & $\checkmark$ & 0.8178 \\
\verb|DeepSeek V3 0324 FSP 3shot[noref]| & $\checkmark$ & 0.8141 \\
\verb|_COMET-22| & & 0.8063 \\
\verb|XCOMET-QE[noref]| & & 0.7996 \\
\verb|BLCOM_1| & & 0.7861 \\
\verb|Qwen2.5 14B FSP 3shot[noref]| & $\checkmark$ & 0.7848 \\
\verb|bright-qe[noref]| & & 0.7818 \\
\verb|*metametrics_mt_mqm_qe_same_source_t[noref]| & & 0.7782 \\
\verb|Qwen2.5 32B FSP 3shot[noref]| & $\checkmark$ & 0.7780 \\
\verb|metametrics_mt_mqm_qe_kendall.seg.s[noref]| & & 0.7739 \\
\verb|_PrismRefMedium| & & 0.7482 \\
\verb|_PrismRefSmall| & & 0.7482 \\
\verb|_sentinel-cand-mqm[noref]| & & 0.7428 \\
\verb|_YiSi-1| & & 0.7274 \\
\verb|damonmonli| & & 0.7267 \\
\verb|_CometKiwi[noref]| & & 0.7170 \\
\verb|*monmonli| & & 0.7139 \\
\verb|_BERTScore| & & 0.7061 \\
\verb|MEE4| & & 0.6914 \\
\verb|_chrF| & & 0.6889 \\
\verb|chrfS| & & 0.6877 \\
\verb|_spBLEU| & & 0.6682 \\
\verb|_BLEU| & & 0.6273 \\
\verb|*BLCOM| & & 0.5507 \\
\verb|*XLsimDA[noref]| & & 0.4634 \\
\verb|XLsimMqm[noref]| & & 0.4634 \\
\verb|_sentinel-ref-mqm| & & 0.0000 \\
\verb|_sentinel-src-mqm[noref]| & & 0.0000 \\

\end{tabular}
\caption{Comparison to the official WMT'24 submissions in terms of system ranking accuracy averaged across three language directions. \texttt{[noref] denotes metrics that are reference free.}}
\label{tab:comparison-to-wmt24-submissions}
\end{table*}

\onecolumn
\section{Prompts}
\label{appendix:prompts}
\subsection{GEMBA prompt variation}
We modify the GEMBA-MQM~\cite{kocmi-federmann-2023-gemba} prompt by changing the output format to JSON for easier parsing and error span explanations, and we present this modified prompt in Figure~\ref{fig:gemba-3-shot-with-explanations}.

\begin{figure*}[h]
\centering
\begin{tcolorbox}[title=GEMBA 3shot prompt with JSON format and error span explanations, breakable=true]
\begin{Verbatim}[breaklines=true, breakanywhere=true, breaksymbolleft={}, fontsize=\scriptsize]
You are an annotator for the quality of machine translation. Your task is to identify errors and assess the quality of the translation using MQM. Based on the source text (in <source></source> tags) and machine translation surrounded (in <translation></translation> tags), identify error types in the translation and classify them. The categories of errors are: accuracy (addition, mistranslation, omission, untranslated text, wrong language), fluency (character encoding, grammar, inconsistency, punctuation, register, spelling), style (awkward), terminology (inappropriate for context, inconsistent use), other. Each error, including omissions or untranslated content, is classified as one of three categories: critical, major, and minor. Critical errors inhibit comprehension of the text. Major errors disrupt the flow, but what the text is trying to say is still understandable. Minor errors are technically errors, but do not disrupt the flow or hinder comprehension. The source text muss be fully covered and any omissions should also be annotated as errors. Please only include errors and no spans that do not contain errors.

Please respond in JSON following this schema:
{
  "type": "object",
  "properties": {
    "errors": {
      "type": "array",
      "items": {
        "type": "object",
        "properties": {
          "error_span": {
            "type": "string",
            "description": "The relevant input span where the error occurred."
          },
          "explanation": {
            "type": "string",
            "description": "A brief explanation of the error and its impact"
          },
          "error_category": {
            "type": "string",
            "enum": ["accuracy", "fluency", "style", "terminology", "other"],
            "description": "The main category of the error"
          },
          "error_type": {
            "type": "string",
            "description": "The specific type of error within the category"
          },
          "severity": {
            "type": "string",
            "enum": ["critical", "major", "minor"],
            "description": "The severity level of the error"
          },
        },
        "required": ["explanation", "error_category", "error_type", "severity"]
      }
    }
  },
  "required": ["errors"]
}
\end{Verbatim}
\begin{Verbatim}[breaklines=true, breakanywhere=true, breaksymbolleft={}, fontsize=\scriptsize, commandchars=\\\{\}]
\colorbox{LightTeal}{<<Example 1>>}
\colorbox{LightTeal}{<<Example 2>>}
\colorbox{LightTeal}{<<Example 3>>}
\end{Verbatim}

\begin{Verbatim}[breaklines=true, breakanywhere=true, breaksymbolleft={}, fontsize=\scriptsize]
Please score the following input
<input>
<source_language>{{ src_lang }}</source_language>
<source>{{ src }}</source>
<target_language>{{ tgt_lang }}</target_language>
<translation>{{ output_seq }}</translation>
</input>

Please respond in JSON without any introduction or explanation. Only the JSON response is required.

MQM:
\end{Verbatim}
\end{tcolorbox}
\caption{The GEMBA 3-shot prompt augmented with error span explanations.  For a clearer presentation, the \colorbox{LightTeal}{three examples} are shown separately in Figures~\ref{fig:gemba-3-shot-with-explanations-example-1}, \ref{fig:gemba-3-shot-with-explanations-example-2}, and~\ref{fig:gemba-3-shot-with-explanations-example-3}.} \label{fig:gemba-3-shot-with-explanations}
\end{figure*}

\begin{figure*}[h]
\centering
\begin{tcolorbox}[title=GEMBA 3shot prompt with JSON format and span explanations Example 1, breakable=true]
\begin{Verbatim}[breaklines=true, breakanywhere=false, breaksymbolleft={}, fontsize=\small]
<input>
<source_language>English</source_language>
<source>I do apologise about this, we must gain permission from the account holder to discuss an order with another person, I apologise if this was done previously, however, I would not be able to discuss this with yourself without the account holders permission.</source>
<target_language>German</target_language>
<translation>Ich entschuldige mich dafür, wir müssen die Erlaubnis einholen, um eine Bestellung mit einer anderen Person zu besprechen. Ich entschuldige mich, falls dies zuvor geschehen wäre, aber ohne die Erlaubnis des Kontoinhabers wäre ich nicht in der Lage, dies mit dir involvement.</translation>
</input>

MQM:
{
  "errors": [
    {
      "error_span": "involvement",
      "explanation": "The word 'involvement' is mistranslated and doesn't fit in the context of the German sentence.",
      "error_category": "accuracy",
      "error_type": "mistranslation",
      "severity": "major"
    },
    {
      "error_span": "",
      "explanation": "The phrase 'the account holder' is omitted in the German translation, losing important information.",
      "error_category": "accuracy",
      "error_type": "omission",
      "severity": "major"
    },
    {
      "error_span": "wäre",
      "explanation": "The use of 'wäre' (subjunctive) is grammatically incorrect in this context. It should be 'bin' (indicative).",
      "error_category": "fluency",
      "error_type": "grammar",
      "severity": "minor"
    },
    {
      "error_span": "dir",
      "explanation": "The use of 'dir' (informal 'you') is inappropriate for the register of this text. It should be 'Ihnen' (formal 'you').",
      "error_category": "fluency",
      "error_type": "register",
      "severity": "minor"
    }
  ]
}
\end{Verbatim}
\end{tcolorbox}
\caption{Example 1 from the GEMBA 3-shot prompt in Jinja format.}
\label{fig:gemba-3-shot-with-explanations-example-1}
\end{figure*}

\begin{figure*}[h]
\centering
\begin{tcolorbox}[title=GEMBA 3shot prompt with JSON format and span explanations Example 2, breakable=true]
\begin{Verbatim}[breaklines=true, breakanywhere=false, breaksymbolleft={}, fontsize=\small]
<input>
<source_language>English</source_language>
<source>Talks have resumed in Vienna to try to revive the nuclear pact, with both sides trying to gauge the prospects of success after the latest exchanges in the stop-start negotiations.</source>
<target_language>Czech</target_language>
<translation>Ve Vídni se ve Vídni obnovily rozhovory o oživení jaderného paktu, přičemže obě partaje se snaží posoudit vyhlídky na úspěch po posledních výměnách v jednáních.</translation>
</input>

MQM:
{
  "errors": [
    {
      "error_span": "ve Vídni se ve Vídni",
      "explanation": "The phrase 've Vídni' (in Vienna) is unnecessarily repeated, adding redundant information.",
      "error_category": "accuracy",
      "error_type": "addition",
      "severity": "major"
    },
    {
      "error_span": "",
      "explanation": "The phrase 'the stop-start' is omitted in the Czech translation, losing the characterization of the negotiations as intermittent.",
      "error_category": "accuracy",
      "error_type": "omission",
      "severity": "major"
    },
    {
      "error_span": "partaje",
      "explanation": "The word 'partaje' (parties) is inappropriate for this context. A more formal term like 'strany' (sides) would be more suitable.",
      "error_category": "terminology",
      "error_type": "inappropriate for context",
      "severity": "minor"
    }
  ]
}
\end{Verbatim}
\end{tcolorbox}
\caption{Example 2 from the GEMBA 3-shot prompt in Jinja format.}
\label{fig:gemba-3-shot-with-explanations-example-2}
\end{figure*}

\begin{figure*}[h]
\centering
\begin{tcolorbox}[title=GEMBA 3shot prompt with JSON format and span explanations Example 3, breakable=true]
\begin{CJK*}{UTF8}{gbsn}
\begin{Verbatim}[breaklines=true, breakanywhere=false, breaksymbolleft={}, fontsize=\small]
<input>
<source_language>Chinese</source_language>
<source>大众点评乌鲁木齐家居商场频道为您提供高铁居然之家地址,电话,营业时间等最新商户信息, 找装修公司,就上大众点评</source>
\end{Verbatim}
\begin{Verbatim}[breaklines=true, breakanywhere=false, breaksymbolleft={}, fontsize=\small]
<target_language>English</target_language>
<translation>Urumqi Home Furnishing Store Channel provides you with the latest business information such as the address, telephone number, business hours, etc., of high-speed rail, and find a decoration company, and go to the reviews.</translation>
</input>

MQM:
{
  "errors": [
    {
      "error_span": "of high-speed rail",
      "explanation": "The phrase 'of high-speed rail' is incorrectly added to the translation. It's not present in the original Chinese text and doesn't make sense in this context.",
      "error_category": "accuracy",
      "error_type": "addition",
      "severity": "critical"
    },
    {
      "error_span": "go to the reviews",
      "explanation": "The phrase 'go to the reviews' is a mistranslation. The original Chinese text refers to using Dianping (a review platform), not simply going to reviews.",
      "error_category": "accuracy",
      "error_type": "mistranslation",
      "severity": "major"
    },
    {
      "error_span": "etc.,",
      "explanation": "The use of 'etc.' is inappropriate in this context, as it introduces vagueness to the translation. The list should be explicit and complete.",
      "error_category": "style",
      "error_type": "awkward",
      "severity": "minor"
    }
  ]
}
\end{Verbatim}
\end{CJK*}
\end{tcolorbox}
\caption{Example 3 from the GEMBA 3-shot prompt in Jinja format.}
\label{fig:gemba-3-shot-with-explanations-example-3}
\end{figure*}

\subsection{Focus Sentence Prompting (FSP)}
\label{prompt:tgtsent}
We present the complete FSP prompt in Figure~\ref{fig:fsp_prompt}.

\include{prompt_fsp.tex}

\subsection{GMICL-5 prompting}
\label{prompt:gmicl-5}
We construct \textit{doc} and \textit{5doc} examples from the following WMT'23 metrics shared task gold annotations:
\begin{itemize}
    \item Doc: \verb|news_guardian.114833:en-de|, System: \verb|NLLB_MBR_BLEU|
    \item Doc: \verb|mastodon_mathewdiekhake.110349821603822000:en-de|, System: \verb|refA|
    \item Doc: \verb|userreview_automotive-2-en_0371449-77:en-de|, System: \verb|AIRC|
    \item Doc: \verb|speech_elitr_minuting-10:en-de|, System: \verb|GPT4-5shot|
    \item Doc: \verb|news_leadership-en.43063:en-de|, System: \verb|ONLINE-M|
    \item Doc: \verb|userreview_luggage-2-en_0553796-30:en-de|, System: \verb|NLLB_MBR_BLEU|
\end{itemize}
The GMICL-5 is presented in Figure~\ref{fig:gmicl-5-prompt}.

\begin{figure*}[h]
\centering
\begin{tcolorbox}[title=GMICL-5 prompt, breakable=true]
\begin{Verbatim}[breaklines=true, breakanywhere=true, breaksymbolleft={}, fontsize=\small, commandchars=\\\{\}]
You are an annotator for the quality of machine translation. Your task is to identify errors and assess the quality of the translation using MQM. Based on the source text (in <source></source> tags) and machine translation surrounded (in <translation></translation> tags), identify error types in the translation and classify them. The categories of errors are: accuracy (addition, mistranslation, omission, untranslated text, wrong language), fluency (character encoding, grammar, inconsistency, punctuation, register, spelling), style (awkward), terminology (inappropriate for context, inconsistent use), other. Each error, including omissions or untranslated content, is classified as one of three categories: critical, major, and minor. Critical errors inhibit comprehension of the text. Major errors disrupt the flow, but what the text is trying to say is still understandable. Minor errors are technically errors, but do not disrupt the flow or hinder comprehension. The source text muss be fully covered and any omissions should also be annotated as errors. Please only include errors and no spans that do not contain errors. 

Please respond in JSON following this schema:
\colorbox{LightTeal}{<<SAME AS THE FSP PROMTP>>}

Here are some examples:
\colorbox{LightGold}{<<FIVE GRANULARITY MATCHED EXAMPLES>>}
\end{Verbatim}
\begin{Verbatim}[breaklines=true, breakanywhere=true, breaksymbolleft={}, fontsize=\small]
Please score the following input
<input>
<source_language>{{ src_lang }}</source_language>
<source>{{ src }}</source>
<target_language>{{ tgt_lang }}</target_language>
<translation>{{ output_seq }}</translation>
</input>

Please respond only in JSON without any introduction. Only the JSON response is required. Unlike the examples you will include error span explanations and a final quality_score.

MQM (with explanation, with quality_score):
\end{Verbatim}

\end{tcolorbox}
\caption{The GMICL-5 prompt in Jinja format. For a clearer presentation, we have omitted the JSON schema for the \colorbox{LightTeal}{MQM error annotations}, which is identical to the one in Figure~\ref{fig:fsp_prompt}, as well as the content of the \colorbox{LightGold}{five documents and their translations}.}
\label{fig:gmicl-5-prompt}

\end{figure*}

\twocolumn

\end{document}

%% file: prompt_fsp.tex
\begin{figure*}[h]
\centering
\begin{tcolorbox}[title=Focus Sentence Prompting (FSP), breakable=true]
\begin{Verbatim}[breaklines=true, breakanywhere=true, breaksymbolleft={}, fontsize=\scriptsize]
You are an annotator for the quality of machine translation. Your task is to identify errors and assess the quality of the translation using MQM. Based on the source text (in <source></source> tags) and machine translation surrounded (in <translation></translation> tags), identify error types in the translation and classify them. The categories of errors are: accuracy (addition, mistranslation, omission, untranslated text, wrong language), fluency (character encoding, grammar, inconsistency, punctuation, register, spelling), style (awkward), terminology (inappropriate for context, inconsistent use), other. Each error, including omissions or untranslated content, is classified as one of three categories: critical, major, and minor. Critical errors inhibit comprehension of the text. Major errors disrupt the flow, but what the text is trying to say is still understandable. Minor errors are technically errors, but do not disrupt the flow or hinder comprehension. The source text muss be fully covered and any omissions should also be annotated as errors. Please only include errors and no spans that do not contain errors. You will be given a full document and its translations, but only score one sentence at a time which is given in <target_segment></target_segment> tags.

Please respond in JSON following this schema:
{
  "type": "object",
  "properties": {
    "errors": {
      "type": "array",
      "items": {
        "type": "object",
        "properties": {
          "error_span": {
            "type": "string",
            "description": "The relevant input span where the error occurred."
          },
          "explanation": {
            "type": "string",
            "description": "A brief explanation of the error and its impact"
          },
          "error_category": {
            "type": "string",
            "enum": ["accuracy", "fluency", "style", "terminology", "other"],
            "description": "The main category of the error"
          },
          "error_type": {
            "type": "string",
            "description": "The specific type of error within the category"
          },
          "severity": {
            "type": "string",
            "enum": ["critical", "major", "minor"],
            "description": "The severity level of the error"
          },
        },
        "required": ["explanation", "error_category", "error_type", "severity"]
      }
    },
    "quality_score": {
      "type": "integer",
      "description": "Overall quality score of the translation. After highlighting all errors, please choose the overall quality score. The quality levels associated with numerical scores: 0: No meaning preserved: Nearly all information is lost in the translation. 33: Some meaning preserved: Some of the meaning is preserved but significant parts are missing. The narrative is hard to follow due to errors. The text may be phrased in an unnatural/awkward way. Grammar may be poor. 66: Most meaning preserved and few grammar mistakes: The translation retains most of the meaning. It may have some grammar mistakes or minor inconsistencies. 100: Perfect meaning and grammar: The meaning and grammar of the translation is completely consistent with the source. The text sounds like native text in the the target language without any awkward phrases. Use any number in the range between 0 and 100 for a fine-grained quality score."
    }
  },
  "required": ["errors", "quality_score"]
}

Please score the following input
<input>
<source_language>{{ src_lang }}</source_language>
<source>{{ src }}</source>
<target_language>{{ tgt_lang }}</target_language>
<translation>{{ output_seq }}</translation>
<target_segment>{{ target_segment }}</target_segment>
</input>

Please respond in JSON without any introduction or explanation. Only the JSON response is required. Use the full document as context while only scoring the translation segment given in <target_segment></target_segment> tags.

MQM:
\end{Verbatim}

\end{tcolorbox}
\caption{The FSP prompt in Jinja format.}
\label{fig:fsp_prompt}
\end{figure*}

%% file: main.bbl
\begin{thebibliography}{24}
\providecommand{\natexlab}[1]{#1}

\bibitem[{Achiam et~al.(2023)Achiam, Adler, Agarwal, Ahmad, Akkaya, Aleman, Almeida, Altenschmidt, Altman, Anadkat et~al.}]{achiam2023gpt}
Josh Achiam, Steven Adler, Sandhini Agarwal, Lama Ahmad, Ilge Akkaya, Florencia~Leoni Aleman, Diogo Almeida, Janko Altenschmidt, Sam Altman, Shyamal Anadkat, et~al. 2023.
\newblock Gpt-4 technical report.
\newblock \emph{arXiv preprint arXiv:2303.08774}.

\bibitem[{Blain et~al.(2023)Blain, Zerva, Rei, Guerreiro, Kanojia, de~Souza, Silva, Vaz, Jingxuan, Azadi et~al.}]{blain2023findings}
Frederic Blain, Chrysoula Zerva, Ricardo Rei, Nuno~M Guerreiro, Diptesh Kanojia, Jos{\'e}~GC de~Souza, Beatriz Silva, T{\^a}nia Vaz, Yan Jingxuan, Fatemeh Azadi, et~al. 2023.
\newblock Findings of the wmt 2023 shared task on quality estimation.
\newblock In \emph{Proceedings of the Eighth Conference on Machine Translation}, pages 629--653.

\bibitem[{Briakou et~al.(2024)Briakou, Luo, Cherry, and Freitag}]{briakou-etal-2024-translating}
Eleftheria Briakou, Jiaming Luo, Colin Cherry, and Markus Freitag. 2024.
\newblock \href {https://doi.org/10.18653/v1/2024.wmt-1.123} {Translating step-by-step: Decomposing the translation process for improved translation quality of long-form texts}.
\newblock In \emph{Proceedings of the Ninth Conference on Machine Translation}, pages 1301--1317, Miami, Florida, USA. Association for Computational Linguistics.

\bibitem[{Conneau et~al.(2020)Conneau, Khandelwal, Goyal, Chaudhary, Wenzek, Guzm{\'a}n, Grave, Ott, Zettlemoyer, and Stoyanov}]{conneau-etal-2020-unsupervised}
Alexis Conneau, Kartikay Khandelwal, Naman Goyal, Vishrav Chaudhary, Guillaume Wenzek, Francisco Guzm{\'a}n, Edouard Grave, Myle Ott, Luke Zettlemoyer, and Veselin Stoyanov. 2020.
\newblock \href {https://doi.org/10.18653/v1/2020.acl-main.747} {Unsupervised cross-lingual representation learning at scale}.
\newblock In \emph{Proceedings of the 58th Annual Meeting of the Association for Computational Linguistics}, pages 8440--8451, Online. Association for Computational Linguistics.

\bibitem[{DeepSeek-AI et~al.(2025)DeepSeek-AI, Liu, Feng, Xue, Wang, Wu, Lu, Zhao, Deng, Zhang, Ruan, Dai, Guo, Yang, Chen, Ji, Li, Lin, Dai, Luo, Hao, Chen, Li, Zhang, Bao, Xu, Wang, Zhang, Ding, Xin, Gao, Li, Qu, Cai, Liang, Guo, Ni, Li, Wang, Chen, Chen, Yuan, Qiu, Li, Song, Dong, Hu, Gao, Guan, Huang, Yu, Wang, Zhang, Xu, Xia, Zhao, Wang, Zhang, Li, Wang, Zhang, Zhang, Tang, Li, Tian, Huang, Wang, Zhang, Wang, Zhu, Chen, Du, Chen, Jin, Ge, Zhang, Pan, Wang, Xu, Zhang, Chen, Li, Lu, Zhou, Chen, Wu, Ye, Ye, Ma, Wang, Zhou, Yu, Zhou, Pan, Wang, Yun, Pei, Sun, Xiao, Zeng, Zhao, An, Liu, Liang, Gao, Yu, Zhang, Li, Jin, Wang, Bi, Liu, Wang, Shen, Chen, Zhang, Chen, Nie, Sun, Wang, Cheng, Liu, Xie, Liu, Yu, Song, Shan, Zhou, Yang, Li, Su, Lin, Li, Wang, Wei, Zhu, Zhang, Xu, Xu, Huang, Li, Zhao, Sun, Li, Wang, Yu, Zheng, Zhang, Shi, Xiong, He, Tang, Piao, Wang, Tan, Ma, Liu, Guo, Wu, Ou, Zhu, Wang, Gong, Zou, He, Zha, Xiong, Ma, Yan, Luo, You, Liu, Zhou, Wu, Ren, Ren, Sha, Fu, Xu, Huang, Zhang, Xie, Zhang, Hao,
  Gou, Ma, Yan, Shao, Xu, Wu, Zhang, Li, Gu, Zhu, Liu, Li, Xie, Song, Gao, and Pan}]{deepseekai2025deepseekv3}
DeepSeek-AI, Aixin Liu, Bei Feng, Bing Xue, Bingxuan Wang, Bochao Wu, Chengda Lu, Chenggang Zhao, Chengqi Deng, Chenyu Zhang, Chong Ruan, Damai Dai, Daya Guo, Dejian Yang, Deli Chen, Dongjie Ji, Erhang Li, Fangyun Lin, Fucong Dai, Fuli Luo, Guangbo Hao, Guanting Chen, Guowei Li, H.~Zhang, Han Bao, Hanwei Xu, Haocheng Wang, Haowei Zhang, Honghui Ding, Huajian Xin, Huazuo Gao, Hui Li, Hui Qu, J.~L. Cai, Jian Liang, Jianzhong Guo, Jiaqi Ni, Jiashi Li, Jiawei Wang, Jin Chen, Jingchang Chen, Jingyang Yuan, Junjie Qiu, Junlong Li, Junxiao Song, Kai Dong, Kai Hu, Kaige Gao, Kang Guan, Kexin Huang, Kuai Yu, Lean Wang, Lecong Zhang, Lei Xu, Leyi Xia, Liang Zhao, Litong Wang, Liyue Zhang, Meng Li, Miaojun Wang, Mingchuan Zhang, Minghua Zhang, Minghui Tang, Mingming Li, Ning Tian, Panpan Huang, Peiyi Wang, Peng Zhang, Qiancheng Wang, Qihao Zhu, Qinyu Chen, Qiushi Du, R.~J. Chen, R.~L. Jin, Ruiqi Ge, Ruisong Zhang, Ruizhe Pan, Runji Wang, Runxin Xu, Ruoyu Zhang, Ruyi Chen, S.~S. Li, Shanghao Lu, Shangyan Zhou, Shanhuang
  Chen, Shaoqing Wu, Shengfeng Ye, Shengfeng Ye, Shirong Ma, Shiyu Wang, Shuang Zhou, Shuiping Yu, Shunfeng Zhou, Shuting Pan, T.~Wang, Tao Yun, Tian Pei, Tianyu Sun, W.~L. Xiao, Wangding Zeng, Wanjia Zhao, Wei An, Wen Liu, Wenfeng Liang, Wenjun Gao, Wenqin Yu, Wentao Zhang, X.~Q. Li, Xiangyue Jin, Xianzu Wang, Xiao Bi, Xiaodong Liu, Xiaohan Wang, Xiaojin Shen, Xiaokang Chen, Xiaokang Zhang, Xiaosha Chen, Xiaotao Nie, Xiaowen Sun, Xiaoxiang Wang, Xin Cheng, Xin Liu, Xin Xie, Xingchao Liu, Xingkai Yu, Xinnan Song, Xinxia Shan, Xinyi Zhou, Xinyu Yang, Xinyuan Li, Xuecheng Su, Xuheng Lin, Y.~K. Li, Y.~Q. Wang, Y.~X. Wei, Y.~X. Zhu, Yang Zhang, Yanhong Xu, Yanhong Xu, Yanping Huang, Yao Li, Yao Zhao, Yaofeng Sun, Yaohui Li, Yaohui Wang, Yi~Yu, Yi~Zheng, Yichao Zhang, Yifan Shi, Yiliang Xiong, Ying He, Ying Tang, Yishi Piao, Yisong Wang, Yixuan Tan, Yiyang Ma, Yiyuan Liu, Yongqiang Guo, Yu~Wu, Yuan Ou, Yuchen Zhu, Yuduan Wang, Yue Gong, Yuheng Zou, Yujia He, Yukun Zha, Yunfan Xiong, Yunxian Ma, Yuting Yan, Yuxiang
  Luo, Yuxiang You, Yuxuan Liu, Yuyang Zhou, Z.~F. Wu, Z.~Z. Ren, Zehui Ren, Zhangli Sha, Zhe Fu, Zhean Xu, Zhen Huang, Zhen Zhang, Zhenda Xie, Zhengyan Zhang, Zhewen Hao, Zhibin Gou, Zhicheng Ma, Zhigang Yan, Zhihong Shao, Zhipeng Xu, Zhiyu Wu, Zhongyu Zhang, Zhuoshu Li, Zihui Gu, Zijia Zhu, Zijun Liu, Zilin Li, Ziwei Xie, Ziyang Song, Ziyi Gao, and Zizheng Pan. 2025.
\newblock \href {https://arxiv.org/abs/2412.19437} {Deepseek-v3 technical report}.
\newblock \emph{Preprint}, arXiv:2412.19437.

\bibitem[{Fernandes et~al.(2023)Fernandes, Deutsch, Finkelstein, Riley, Martins, Neubig, Garg, Clark, Freitag, and Firat}]{fernandes-etal-2023-devil}
Patrick Fernandes, Daniel Deutsch, Mara Finkelstein, Parker Riley, Andr{\'e} Martins, Graham Neubig, Ankush Garg, Jonathan Clark, Markus Freitag, and Orhan Firat. 2023.
\newblock \href {https://doi.org/10.18653/v1/2023.wmt-1.100} {The devil is in the errors: Leveraging large language models for fine-grained machine translation evaluation}.
\newblock In \emph{Proceedings of the Eighth Conference on Machine Translation}, pages 1066--1083, Singapore. Association for Computational Linguistics.

\bibitem[{Freitag et~al.(2024)Freitag, Mathur, Deutsch, Lo, Avramidis, Rei, Thompson, Blain, Kocmi, Wang, Adelani, Buchicchio, Zerva, and Lavie}]{freitag-etal-2024-llms}
Markus Freitag, Nitika Mathur, Daniel Deutsch, Chi-Kiu Lo, Eleftherios Avramidis, Ricardo Rei, Brian Thompson, Frederic Blain, Tom Kocmi, Jiayi Wang, David~Ifeoluwa Adelani, Marianna Buchicchio, Chrysoula Zerva, and Alon Lavie. 2024.
\newblock \href {https://doi.org/10.18653/v1/2024.wmt-1.2} {Are {LLM}s breaking {MT} metrics? results of the {WMT}24 metrics shared task}.
\newblock In \emph{Proceedings of the Ninth Conference on Machine Translation}, pages 47--81, Miami, Florida, USA. Association for Computational Linguistics.

\bibitem[{Freitag et~al.(2023)Freitag, Mathur, Lo, Avramidis, Rei, Thompson, Kocmi, Blain, Deutsch, Stewart, Zerva, Castilho, Lavie, and Foster}]{freitag-etal-2023-results}
Markus Freitag, Nitika Mathur, Chi-kiu Lo, Eleftherios Avramidis, Ricardo Rei, Brian Thompson, Tom Kocmi, Frederic Blain, Daniel Deutsch, Craig Stewart, Chrysoula Zerva, Sheila Castilho, Alon Lavie, and George Foster. 2023.
\newblock \href {https://doi.org/10.18653/v1/2023.wmt-1.51} {Results of {WMT}23 metrics shared task: Metrics might be guilty but references are not innocent}.
\newblock In \emph{Proceedings of the Eighth Conference on Machine Translation}, pages 578--628, Singapore. Association for Computational Linguistics.

\bibitem[{Guerreiro et~al.(2024)Guerreiro, Rei, Stigt, Coheur, Colombo, and Martins}]{guerreiro-etal-2024-xcomet}
Nuno~M. Guerreiro, Ricardo Rei, Daan~van Stigt, Luisa Coheur, Pierre Colombo, and Andr{\'e} F.~T. Martins. 2024.
\newblock \href {https://doi.org/10.1162/tacl_a_00683} {xcomet: Transparent machine translation evaluation through fine-grained error detection}.
\newblock \emph{Transactions of the Association for Computational Linguistics}, 12:979--995.

\bibitem[{Kocmi and Federmann(2023{\natexlab{a}})}]{kocmi-federmann-2023-gemba}
Tom Kocmi and Christian Federmann. 2023{\natexlab{a}}.
\newblock \href {https://doi.org/10.18653/v1/2023.wmt-1.64} {{GEMBA}-{MQM}: Detecting translation quality error spans with {GPT}-4}.
\newblock In \emph{Proceedings of the Eighth Conference on Machine Translation}, pages 768--775, Singapore. Association for Computational Linguistics.

\bibitem[{Kocmi and Federmann(2023{\natexlab{b}})}]{kocmi-federmann-2023-large}
Tom Kocmi and Christian Federmann. 2023{\natexlab{b}}.
\newblock \href {https://aclanthology.org/2023.eamt-1.19/} {Large language models are state-of-the-art evaluators of translation quality}.
\newblock In \emph{Proceedings of the 24th Annual Conference of the European Association for Machine Translation}, pages 193--203, Tampere, Finland. European Association for Machine Translation.

\bibitem[{Kocmi et~al.(2021)Kocmi, Federmann, Grundkiewicz, Junczys-Dowmunt, Matsushita, and Menezes}]{kocmi-etal-2021-ship}
Tom Kocmi, Christian Federmann, Roman Grundkiewicz, Marcin Junczys-Dowmunt, Hitokazu Matsushita, and Arul Menezes. 2021.
\newblock \href {https://aclanthology.org/2021.wmt-1.57/} {To ship or not to ship: An extensive evaluation of automatic metrics for machine translation}.
\newblock In \emph{Proceedings of the Sixth Conference on Machine Translation}, pages 478--494, Online. Association for Computational Linguistics.

\bibitem[{Levy et~al.(2024)Levy, Jacoby, and Goldberg}]{levy-etal-2024-task}
Mosh Levy, Alon Jacoby, and Yoav Goldberg. 2024.
\newblock \href {https://doi.org/10.18653/v1/2024.acl-long.818} {Same task, more tokens: the impact of input length on the reasoning performance of large language models}.
\newblock In \emph{Proceedings of the 62nd Annual Meeting of the Association for Computational Linguistics (Volume 1: Long Papers)}, pages 15339--15353, Bangkok, Thailand. Association for Computational Linguistics.

\bibitem[{Lommel et~al.(2014)Lommel, Uszkoreit, and Burchardt}]{lommel2014multidimensional}
Arle Lommel, Hans Uszkoreit, and Aljoscha Burchardt. 2014.
\newblock Multidimensional quality metrics (mqm): A framework for declaring and describing translation quality metrics.
\newblock \emph{Tradum{\`a}tica}, (12):0455--463.

\bibitem[{Maruf et~al.(2022)Maruf, Saleh, and Haffari}]{maruf-docmt-19}
Sameen Maruf, Fahimeh Saleh, and Gholamreza Haffari. 2022.
\newblock \href {https://doi.org/10.1145/3441691} {A survey on document-level neural machine translation: Methods and evaluation}.
\newblock \emph{{ACM} Comput. Surv.}, 54(2):45:1--45:36.

\bibitem[{Raunak et~al.(2024)Raunak, Kocmi, and Post}]{raunak2023slide}
Vikas Raunak, Tom Kocmi, and Matt Post. 2024.
\newblock \href {https://doi.org/10.18653/v1/2024.naacl-short.18} {{SLIDE}: Reference-free evaluation for machine translation using a sliding document window}.
\newblock In \emph{Proceedings of the 2024 Conference of the North American Chapter of the Association for Computational Linguistics: Human Language Technologies (Volume 2: Short Papers)}, pages 205--211, Mexico City, Mexico. Association for Computational Linguistics.

\bibitem[{Thompson et~al.(2024)Thompson, Mathur, Deutsch, and Khayrallah}]{thompson-etal-2024-improving}
Brian Thompson, Nitika Mathur, Daniel Deutsch, and Huda Khayrallah. 2024.
\newblock \href {https://doi.org/10.18653/v1/2024.wmt-1.118} {Improving statistical significance in human evaluation of automatic metrics via soft pairwise accuracy}.
\newblock In \emph{Proceedings of the Ninth Conference on Machine Translation}, pages 1222--1234, Miami, Florida, USA. Association for Computational Linguistics.

\bibitem[{Treviso et~al.(2024)Treviso, Guerreiro, Agrawal, Rei, Pombal, Vaz, Wu, Silva, Stigt, and Martins}]{treviso-etal-2024-xtower}
Marcos~V Treviso, Nuno~M Guerreiro, Sweta Agrawal, Ricardo Rei, Jos{\'e} Pombal, Tania Vaz, Helena Wu, Beatriz Silva, Daan~Van Stigt, and Andre Martins. 2024.
\newblock \href {https://doi.org/10.18653/v1/2024.findings-emnlp.892} {x{T}ower: A multilingual {LLM} for explaining and correcting translation errors}.
\newblock In \emph{Findings of the Association for Computational Linguistics: EMNLP 2024}, pages 15222--15239, Miami, Florida, USA. Association for Computational Linguistics.

\bibitem[{Vernikos et~al.(2022)Vernikos, Thompson, Mathur, and Federico}]{vernikos-etal-2022-embarrassingly}
Giorgos Vernikos, Brian Thompson, Prashant Mathur, and Marcello Federico. 2022.
\newblock \href {https://aclanthology.org/2022.wmt-1.6/} {Embarrassingly easy document-level {MT} metrics: How to convert any pretrained metric into a document-level metric}.
\newblock In \emph{Proceedings of the Seventh Conference on Machine Translation (WMT)}, pages 118--128, Abu Dhabi, United Arab Emirates (Hybrid). Association for Computational Linguistics.

\bibitem[{Wei et~al.(2022)Wei, Wang, Schuurmans, Bosma, ichter, Xia, Chi, Le, and Zhou}]{neurips-2022-cot}
Jason Wei, Xuezhi Wang, Dale Schuurmans, Maarten Bosma, brian ichter, Fei Xia, Ed~Chi, Quoc~V Le, and Denny Zhou. 2022.
\newblock \href {https://proceedings.neurips.cc/paper_files/paper/2022/file/9d5609613524ecf4f15af0f7b31abca4-Paper-Conference.pdf} {Chain-of-thought prompting elicits reasoning in large language models}.
\newblock In \emph{Advances in Neural Information Processing Systems}, volume~35, pages 24824--24837. Curran Associates, Inc.

\bibitem[{Wu et~al.(2024)Wu, Vu, Qu, Foster, and Haffari}]{wu2024adapting}
Minghao Wu, Thuy-Trang Vu, Lizhen Qu, George Foster, and Gholamreza Haffari. 2024.
\newblock Adapting large language models for document-level machine translation.
\newblock \emph{arXiv preprint arXiv:2401.06468}.

\bibitem[{Yang et~al.(2025)Yang, Yang, Zhang, Hui, Zheng, Yu, Li, Liu, Huang, Wei, Lin, Yang, Tu, Zhang, Yang, Yang, Zhou, Lin, Dang, Lu, Bao, Yang, Yu, Li, Xue, Zhang, Zhu, Men, Lin, Li, Tang, Xia, Ren, Ren, Fan, Su, Zhang, Wan, Liu, Cui, Zhang, and Qiu}]{qwen2025qwen25}
An~Yang, Baosong Yang, Beichen Zhang, Binyuan Hui, Bo~Zheng, Bowen Yu, Chengyuan Li, Dayiheng Liu, Fei Huang, Haoran Wei, Huan Lin, Jian Yang, Jianhong Tu, Jianwei Zhang, Jianxin Yang, Jiaxi Yang, Jingren Zhou, Junyang Lin, Kai Dang, Keming Lu, Keqin Bao, Kexin Yang, Le~Yu, Mei Li, Mingfeng Xue, Pei Zhang, Qin Zhu, Rui Men, Runji Lin, Tianhao Li, Tianyi Tang, Tingyu Xia, Xingzhang Ren, Xuancheng Ren, Yang Fan, Yang Su, Yichang Zhang, Yu~Wan, Yuqiong Liu, Zeyu Cui, Zhenru Zhang, and Zihan Qiu. 2025.
\newblock \href {https://arxiv.org/abs/2412.15115} {Qwen2.5 technical report}.
\newblock \emph{Preprint}, arXiv:2412.15115.

\bibitem[{Zheng et~al.(2023)Zheng, Chiang, Sheng, Zhuang, Wu, Zhuang, Lin, Li, Li, Xing, Zhang, Gonzalez, and Stoica}]{zheng2023judging}
Lianmin Zheng, Wei-Lin Chiang, Ying Sheng, Siyuan Zhuang, Zhanghao Wu, Yonghao Zhuang, Zi~Lin, Zhuohan Li, Dacheng Li, Eric.~P Xing, Hao Zhang, Joseph~E. Gonzalez, and Ion Stoica. 2023.
\newblock \href {https://arxiv.org/abs/2306.05685} {Judging llm-as-a-judge with mt-bench and chatbot arena}.
\newblock \emph{Preprint}, arXiv:2306.05685.

\bibitem[{Zheng et~al.(2024)Zheng, Yin, Xie, Sun, Huang, Yu, Cao, Kozyrakis, Stoica, Gonzalez et~al.}]{zheng2024sglang}
Lianmin Zheng, Liangsheng Yin, Zhiqiang Xie, Chuyue~Livia Sun, Jeff Huang, Cody~Hao Yu, Shiyi Cao, Christos Kozyrakis, Ion Stoica, Joseph~E Gonzalez, et~al. 2024.
\newblock Sglang: Efficient execution of structured language model programs.
\newblock \emph{Advances in Neural Information Processing Systems}, 37:62557--62583.

\end{thebibliography}
